\pdfoutput=1

\documentclass[11pt]{article}
\usepackage[table,xcdraw]{xcolor}
\usepackage[preprint]{acl}

\usepackage{times}
\usepackage{latexsym}
\usepackage[T1]{fontenc}
\usepackage[utf8]{inputenc}
\usepackage{microtype}
\usepackage{inconsolata}
\usepackage{graphicx}
\usepackage{amsmath}
\usepackage{amssymb}
\usepackage{amsfonts}
\usepackage{dsfont}
\usepackage{booktabs}
\usepackage{multirow}
\usepackage{pifont}
\usepackage{mwe}
\usepackage{subcaption}
\usepackage{algorithm}
\usepackage{algpseudocode}
\usepackage{tcolorbox}

\definecolor{JSViolet}{RGB}{71,15,244}
\definecolor{JSRed}{RGB}{205,44,78}
\newcommand{\cmark}{\textcolor{JSViolet}{\ding{51}}}
\newcommand{\xmark}{\textcolor{JSRed}{\ding{55}}}

\title{Think Clearly: Improving Reasoning via Redundant Token Pruning
}

\author{
Daewon Choi$^{1}$,
\,
Jimin Lee$^{2, \dagger}$,
\,
Jihoon Tack$^{1}$,
\,
Woomin Song$^{1, 3, \ddagger}$,
\,
\\
\textbf{Saket Dingliwal}$^{3}$,
\textbf{Sai Muralidhar Jayanthi}$^{3}$,
\textbf{Bhavana Ganesh}$^{3}$,
\textbf{Jinwoo Shin}$^{1}$,
\\
\textbf{Aram Galstyan}$^{3}$,
\textbf{Sravan Babu Bodapati}$^{3}$,
\\
$^{1}$KAIST
$^{2}$Korea University
$^{3}$Amazon AGI}

\usepackage[symbol]{footmisc}

\begin{document}
\maketitle
\footnotetext{$^{\dagger}$ Work done during an internship at KAIST. \ignorespaces$^{\ddagger}$ Work done during an internship at Amazon.
}
\begin{abstract}
Recent large language models have shown promising capabilities in long-form reasoning, following structured chains of thought before arriving at a final answer. 
However, we observe that these reasoning paths tend to include substantial redundancy; analyzing attention patterns reveals that attention scores are widely scattered, particularly incorrect answers exhibit greater attention sparsity. In this paper, we demonstrate that deliberately removing this redundancy in the reasoning process significantly improves performance through clear thinking, i.e., removing distraction. Specifically, we systematically identify reasoning redundancy by measuring token-level attention scores to a special end-of-thinking token, which is appended to an explicit instruction inserted to conclude each intermediate reasoning step.
Furthermore, we propose structure-aware pruning that prioritizes removing tokens in low-contributing reasoning chunks over individual tokens. After evicting redundant tokens, we remove the injected end-of-thinking instruction, then resume the reasoning generation. We demonstrate that our method significantly improves overall accuracy across reasoning-intensive benchmarks without any training involved. In particular, our method shows strong performance on challenging mathematical competition benchmarks such as AIME and AMC, where reasoning redundancy is more prevalent.
\end{abstract}
\section{Introduction}\label{sec:intro}

Large language models (LLMs) have demonstrated remarkable progress in complex reasoning tasks \citep{wei2022chain,zelikman2022star}, including mathematical problem solving \citep{shao2024deepseekmath}, multi-hop question answering \citep{chen2019multi}, and long-form instruction following \citep{bai2024longalign}. 
This success is often attributed to the emergence of structured reasoning chains, generating sequences of intermediate thoughts that gradually lead to a final answer \citep{kojima2022large,hao2024training}.
These reasoning chains allow models to break down complex problems into smaller, more manageable subproblems, mimicking the step-by-step cognitive strategies employed in human reasoning \citep{prystawski2023think}. 

In particular, recent reasoning models are trained to verbalize their internal thoughts, effectively leveraging the language abilities of pre-trained models \citep{guo2025deepseekr1, jaech2024o1}. Additionally, this verbalization offers a key advantage: it allows users to monitor and analyze—or even intervene in—the model’s thought process during generation \citep{baker2025monitoring,wu2025effectively}. 

\begin{figure*}[t]
    \centering
    \begin{subfigure}{0.583\textwidth}
        \includegraphics[width=\linewidth]{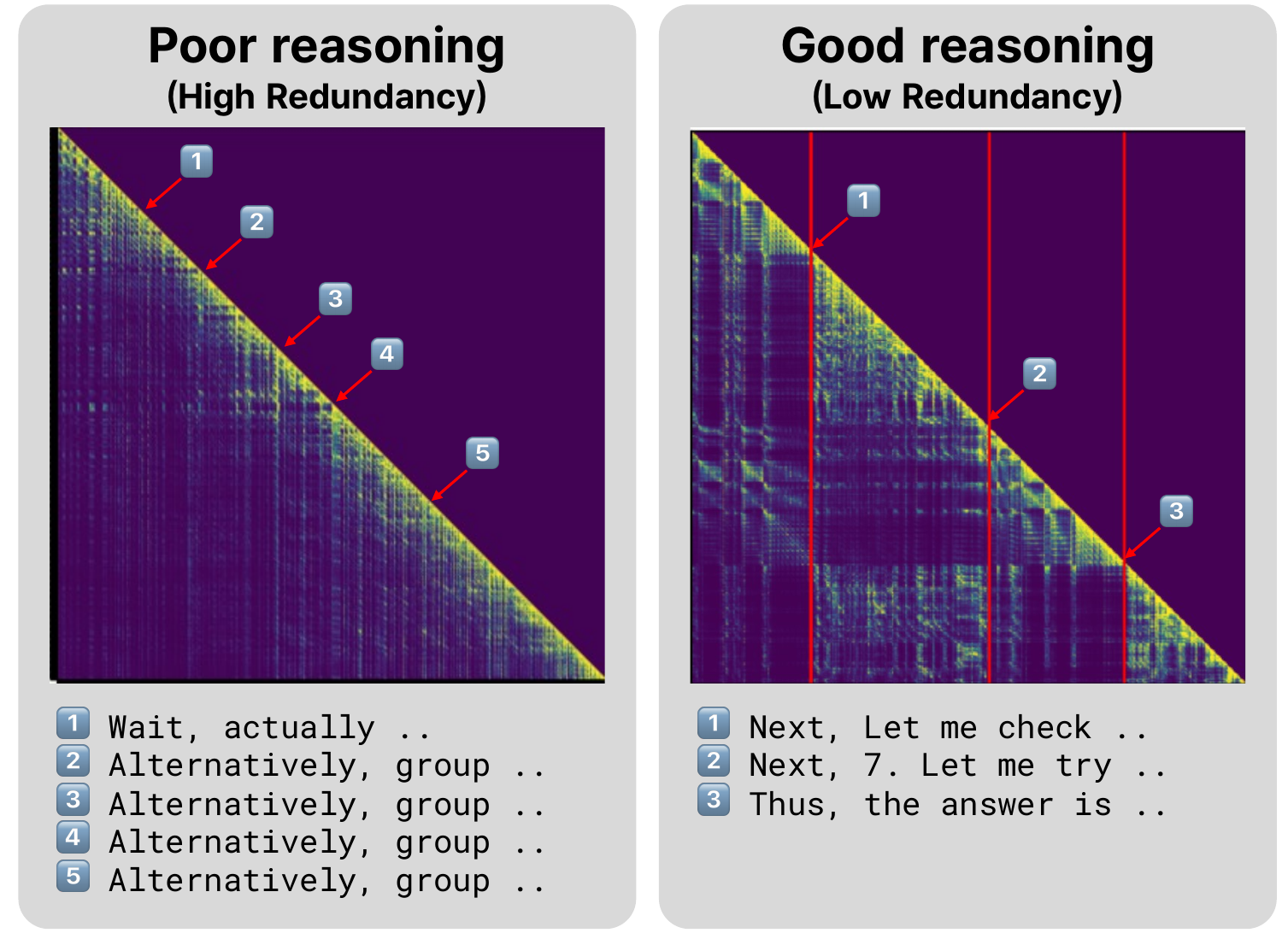}
        \caption{Attention map and corresponding text}
        \label{fig:attention-scatter}
    \end{subfigure}
    \hfill
    \begin{subfigure}{0.4\textwidth}
       \includegraphics[width=\linewidth]{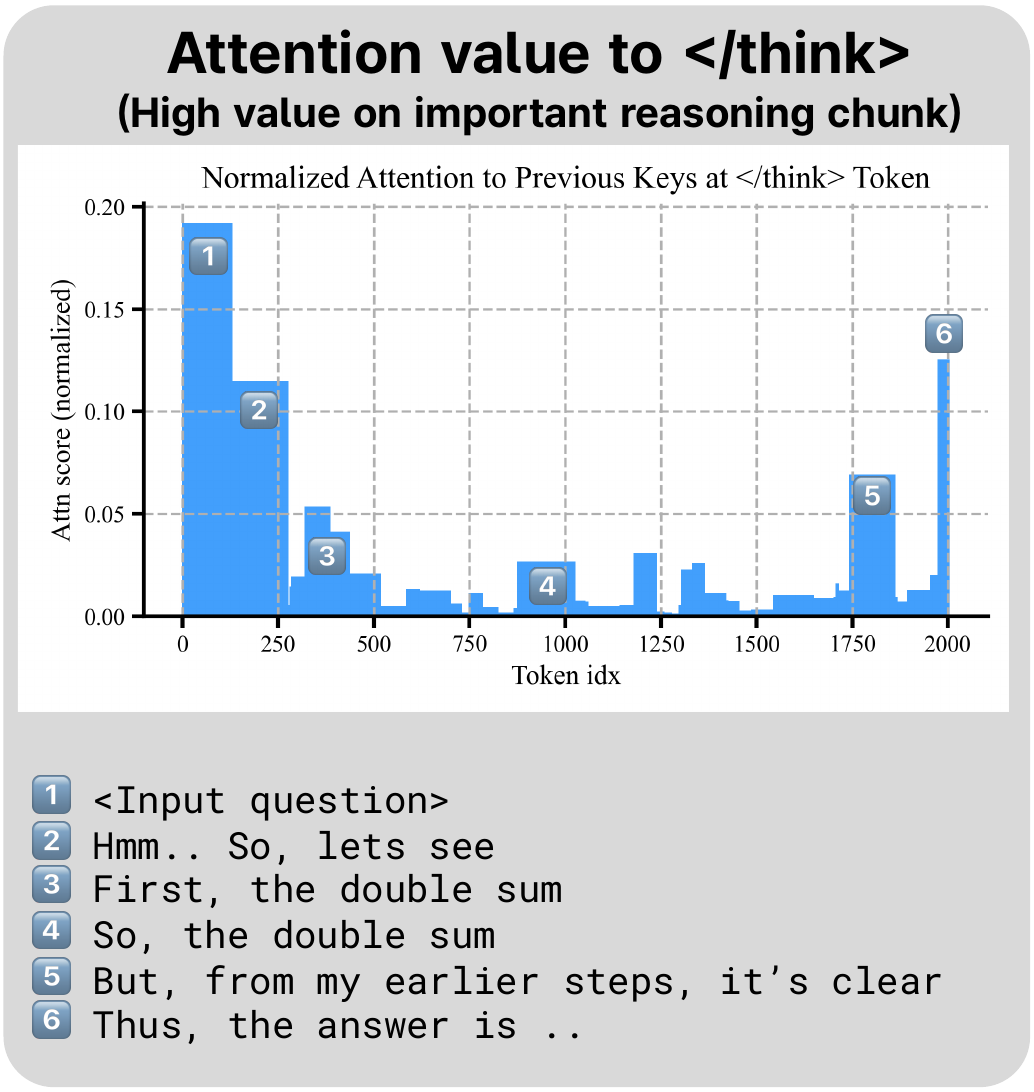}
        \caption{Attention score on important chunks}
        \label{fig:summarization-ablation}
    \end{subfigure}
    \vspace{-0.07in}
    \caption{\textbf{Not all tokens matter for reasoning.} 
    We visualize and analyze the attention map of the output sequence.
    (a) Attention maps when the model fails to produce the correct answer (i.e., poor reasoning) and when it succeeds (i.e., good reasoning). Poor reasoning leads to highly redundant attention patterns.
    (b) Attention scores associated with the end-of-thinking token </think>. The histogram shows that </think> attends to key reasoning chunks that contain crucial information for deriving the final answer.
    We use Qwen2.5-7B distilled from DeepSeek-R1 on a subset of the MATH-500 dataset.
    }
    \vspace{-0.13in}
    \label{fig:reasoning_token_analysis}
\end{figure*}

In this paper, we found a somewhat interesting observation by monitoring this internal thought process of reasoning LLMs: \textit{reasoning chains often consist of significant redundancy}. 
Specifically, the model generates intermediate reasoning steps that tend to be repetitive, verbose, or include speculative detours that do not ultimately contribute to the final answer. 
Such redundancies are also observed by analyzing the attention patterns, where attention distributions during reasoning typically consist of sparse patterns. 
This is especially problematic as LLM can be easily distracted by irrelevant or redundant context \citep{shi2023large}, and this tendency is particularly evident when incorrect answers are generated (see Fig.~\ref{fig:reasoning_token_analysis}). 

More intriguingly, we observe reasoning chunks that receive consistently low attention from subsequent tokens, suggesting that the model briefly explores these misleading paths but eventually abandons them, leaving behind redundant traces in the generated sequence.
This raises a key question:
\begin{center}
\emph{Can we improve the performance 
by identifying and removing redundant tokens on-the-fly during the reasoning process?}
\end{center}

To this end, we propose a simple yet effective test-time token pruning method that removes redundant reasoning tokens. 
The core idea is to dynamically eliminate redundant tokens during generation, thereby enabling the model to preserve only the most critical reasoning steps necessary for reaching the correct answer.
Specifically, we propose two components: (i) identifying redundant tokens by measuring their contribution to a summarization-inducing end-of-thinking token, and (ii) structure-aware pruning that prioritizes removing low-contributing reasoning chunks rather than individual tokens.
Motivated by the observation that redundant tokens tend to receive low attention from the token that concludes the reasoning step, we inject an explicit instruction that prompts the model to summarize and terminate the current thought process, enabling redundancy measurement at intermediate stages.\footnote{
We perform such an additional (summarization) prompting every 200 token-generation and observe that it introduces marginal overhead in inference speed.
}
Rather than removing tokens in isolation, we first detect reasoning chunks that are unlikely to contribute to the final answer (i.e., misleading paths), and prune tokens within those chunks.
Once pruning is completed, we resume the generation by removing the injected end-of-thinking instruction.

We conduct a comprehensive set of experiments to evaluate our token pruning scheme, focusing on reasoning-heavy scenarios across a wide range of tasks and models.
Our results demonstrate that this simple and lightweight inference-time approach can significantly improve reasoning performance.
In particular, our method yields significant gains on challenging mathematics competition benchmarks such as AIME and AMC, where the model tends to generate more redundant reasoning steps, making our pruning approach especially effective.
For instance, our method improves the original model's accuracy from 75.0\% to 82.5\% on AMC2023 \citep{aimo2023aimo}, while reducing KV cache memory usage by 10.3\% on DeepSeek-R1-Distill-Qwen-7B \citep{Yang2024Qwen25TR}.
\section{Related Work}
\label{sec:related-works}

We provide a comprehensive review of related works on reasoning (focusing on the long chain-of-thought literature) and token pruning and compression frameworks.

\paragraph{Long chain-of-thought.}
Recent Large Reasoning Models (LRMs) \cite{guo2025deepseekr1, jaech2024o1} have increasingly adopted explicit chain-of-thought (CoT) reasoning to enhance performance on complex tasks such as math \citep{shao2024deepseekmath}, science \citep{qwen2024qwq}, and symbolic reasoning \citep{xu2024faithful}. Instead of producing direct answers, these models generate multi-step intermediate reasoning traces, allowing them to break down complex problems into smaller, more manageable steps \citep{kojima2022large,prystawski2023think}. This long-form reasoning improves both accuracy and robustness \citep{wang2024offline,guan2024deliberative}, as it provides opportunities for self-correction \citep{kumar2025score}, verification \citep{lee2025revise}, and intermediate supervision during training or inference \citep{wu2025effectively}. A common pattern involves separating the reasoning phase from the answer phase, either through special tokens or by internal abstraction, where the reasoning is hidden and only summarized \citep{hammoud2025beyond}. Some models expose the entire reasoning trace to the user to increase interpretability and transparency \citep{baker2025monitoring}, while others keep it latent to reduce vulnerability to prompt manipulation or over-reliance \citep{hao2024training}. This shift toward structured, multi-step reasoning has been central to the recent progress of reasoning-focused LLMs, enabling strong generalization to diverse domains, from mathematics to program synthesis \citep{gao2023pal}. In this paper, we propose an efficient yet effective test-time reasoning method by pruning redundant tokens, allowing the model to focus on more relevant and critical steps while maintaining or even improving task performance.

\paragraph{Token pruning and compression.} 
As the context length and generated sequences of large language models (LLMs) increase, the memory cost of self-attention becomes a significant bottleneck \citep{dao2022flashattention}. 
In particular, the key-value (KV) cache, which stores past activations for each generated token, grows linearly with the sequence length. This growth introduces a bottleneck during inference, especially in resource-constrained or real-time settings.
To address this, several existing works have explored strategies to identify and evict redundant tokens in KV cache \citep{li2024snapkv}.
\citet{xiao2023streamingllm} observe that attention distributions often exhibit strong focus on initial tokens, and show that retaining only the initial and most recent tokens is sufficient to preserve performance. 
Other approaches leverage attention-based metrics to guide KV eviction \citep{chen2024arkvale}. For example, \citet{zhang2023h2o} proposes accumulating attention scores over decoding steps and using them as token importance indicators. \citet{oren2024tova} evict the token with the lowest attention score at each decoding step. \citet{yang2024pyramidinfer} observe that the number of crucial tokens varies across layers, motivating a layer-wise compression strategy. 
While these methods primarily aim to improve inference efficiency, our approach focuses on eviting redundant reasoning tokens to improve the performance (compared to the full KV cache). 
Nevertheless, we demonstrate the potential of using our method as a reasoning compression scheme, achieving competitive performance under memory constraints and often exceeding recent token eviction and compression baselines.

\section{Not All Tokens Matter for Reasoning}\label{sec:method:obs}

In this section, we investigate whether reasoning models truly require all previously generated tokens to reach a correct final answer. 
To this end, we focus on recent reasoning LLMs output $\mathcal{T}_{\text{output}}$, which consists of multi-step intermediate reasoning traces $\mathcal{T}_{\text{reason}}$ followed by final answer $\mathcal{T}_{\text{answer}}$ with special delimiters $\text{<think>... </think>}$.  

\paragraph{Existence of redundant reasoning tokens.}
To identify which reasoning tokens are important for reaching the answer, we visualize the attention maps of two samples: (i) a sample that fails to answer the given question, and (ii) a sample that successfully reaches the end-of-thinking token to produce the correct answer. As shown in Fig.~\ref{fig:attention-scatter}, the first case includes redundant text (e.g., repeated attempts to rethink the process using phrases like ``Alternatively''), whereas the second case exhibits a clear reasoning trajectory. Interestingly, the attention maps reflect this behavior; in the first case, the attention is highly sparse due to redundancy in reasoning, while in the second case, the model attends more frequently to previously generated tokens and demonstrates a more global structure. These results highlight the potential of using attention scores to identify redundant reasoning steps.

\paragraph{Attention score to </think>.} 
To quantify token importance more systematically, we analyze the attention scores directed to the special token </think>, which marks the end of the reasoning process.
As shown in Fig.~\ref{fig:summarization-ablation}, reasoning tokens that contribute to the final answer tend to have high attention scores to </think>. For example, the </think> token frequently attends to sentences or chunks that initiate the reasoning process or summarize key conclusions.
Interestingly, redundant tokens often appear in contiguous chunks rather than in isolation, suggesting that the </think> token selectively attends to informative reasoning segments while ignoring irrelevant parts.

Based on this observation, we design a systematic token pruning strategy that leverages the end-of-thinking token </think> and the chunked structure of the reasoning process to identify and remove redundant reasoning tokens.

\section{Improving Reasoning with Redundant Token Pruning}
\label{sec:method}

In this section, we propose a novel KV cache eviction policy that can improve reasoning models' effectiveness and efficiency by removing redundant tokens. Specifically, we suggest a novel scoring function driven by self-summarization to identify redundant tokens (in Section \ref{sec:method:self-summ}), and introduce a stepwise eviction policy that aggressively removes KV cache in the redundant reasoning step (in Section \ref{sec:method:stepwise-evict}). The overall procedure is illustrated in Algorithm~\ref{alg:eviction}.
\begin{algorithm}[t]
\caption{Redundant Token Eviction via Self-summarization}
\label{alg:eviction}
\begin{algorithmic}[1]
\Require Reasoning tokens $T = \{t_1, \dots, t_L\}$, eviction budget $k$, layers $\ell \in [1, L]$, heads $h \in [1, H]$
\Ensure Set of $k$ tokens to evict from KV cache

\State Inject summarization prompt $\mathcal{T}_\text{summ}$ into the input
\State Forward the input with the trigger token \text{</think> to the model}

\For{each layer $\ell$ and head $h$}
    \For{each token $t \in T$}
        \State $s_t^{(\ell, h)} \gets \alpha^{(\ell, h)}_{\text{</think>} \rightarrow t}$
    \EndFor
    \State Segment $\mathcal{T}_{\text{reason}}$ into steps $\{r_1, \dots, r_N\}$
    \For{each step $r_i$}
        \State $c_{r_i}^{(\ell)} \gets \frac{1}{|H \cdot r_i|} \sum_{h \in H} \sum_{t \in r_i} s_t^{(\ell, h)}$
    \EndFor
    \State Sort steps $\tilde{r}_1^{(\ell)}, \dots, \tilde{r}_N^{(\ell)}$ in ascending order \\
    \quad\, of $c_{r_i}^{(\ell)}$
    \State $k_{\text{rem}} \gets k$
    \For{each sorted reasoning step $\tilde{r}_i^{(\ell)}$}
        \State $e_{\tilde{r}_i}^{(\ell)} \gets \min\left(|\tilde{r}_i^{(\ell)}|,\; k_{\text{rem}}\right)$
        \State $k_{\text{rem}} \gets k_{\text{rem}} - e_{\tilde{r}_i}^{(\ell)}$
        \State Evict $e_{\tilde{r}_i}^{(\ell)}$ tokens with lowest $s_t^{(\ell, h)}$ in \\
        \quad\quad\quad$\tilde{r}_i$ per head $h$ in layer $\ell$
        \If{$k_{\text{rem}} = 0$}
            \State \textbf{break}
        \EndIf
    \EndFor
\EndFor
\end{algorithmic}
\end{algorithm}

\subsection{Identifying redundant tokens via self-summarization}\label{sec:method:self-summ}
As shown in Fig.~\ref{fig:reasoning_token_analysis}, the end-of-thinking token $\text{</think>}$ serves as a crucial cue for identifying important reasoning tokens. 
Based on this, we propose a novel scoring function for identifying redundant tokens in the reasoning trace during the intermediate decoding so that one can only preserve important tokens. 
Specifically, we 
leverage a short summarization prompt ending with \text{</think>}, prompting the model to briefly summarize its own reasoning. 
This forces the LLM to end the thinking process, thereby effectively localizing the essential part inside the reasoning trace.

\paragraph{Use of summarization prompts.} 
During the intermediate step of the decoding, we periodically trigger the model to summarize and answer the question at every fixed interval.
Here, to evaluate the redundancy of tokens during reasoning, our key idea is to forward the reasoning model with a short summarization prompt $\mathcal{T_{\text{summ.}}}$ which is constructed as follows:  
\begin{quote}
\texttt{``Time is up. Given the time I've spent and the approaches I've tried, I should stop thinking and now write summarization in one sentence.</think>''}  
\end{quote}
Especially, the prompt is designed to explicitly shift the model from reasoning to summarization, making the </think> token to capture informative tokens in the reasoning trace without generating explicit summarization. 

\paragraph{Token importance score.}  
To quantify the importance of each token, we accumulate attention weights assigned to previous tokens given the summarization prompt $\mathcal{T_{\text{summ.}}}$. Specifically, for each token $t$ in the current reasoning trace, we define its importance score $s_t^{(\ell, h)}$ at layer $\ell$ and head $h$ by aggregating attention values by injecting the summarization prompt with $\text{</think>}$:
\begin{equation}
s_t^{(\ell, h)} = \alpha^{(\ell,h)}_{\text{</think>}\rightarrow t},
\end{equation}
where $\alpha^{(\ell,h)}_{\text{</think>} \rightarrow t}$ denotes the attention weight from the $\text{</think>}$ in summarization tokens $\mathcal{ T}_\texttt{summ.}$ at layer $\ell$ and head $h$. This score reflects how much each token contributes to the final summarization, as perceived by the model. Since the scores are computed separately for each layer and attention head, pruning decisions are made independently at each level, enabling fine-grained control over which tokens are retained.

\subsection{Step-aware eviction with hierarchical budget allocation.}\label{sec:method:stepwise-evict}
We now present our eviction policy under a fixed token eviction budget $k$. Motivated by our observation that reasoning traces often contain redundant steps, we aim to remove tokens from such steps while preserving essential ones. To this end, we first segment the reasoning trace into semantically coherent steps and then allocate the eviction budget hierarchically across these steps based on the importance score.

\paragraph{Aggregating importance score per reasoning step.} Following a previous work \cite{hammoud2025subthought}, we first divide the reasoning trace into intermediate steps, and each steps consists of a consecutive set of tokens with logical continuity in reasoning (See Appendix~\ref{app:imp-detail} for detail). Given importance score $s_t^{(\ell, h)}$, we compute step score $c_r^{(\ell)}$ by taking the mean over all tokens $t$ within the same reasoning step $r$
across head $h$:
\begin{equation}
c_r^{(\ell)} = \frac{1}{|H \cdot r|}\sum_{h \in H}\sum_{t \in r} s_t^{(\ell, h)},
\end{equation}
where $|H|$ is the number of heads and $|r|$ is length of reasoning step.

\paragraph{Hierarchical eviction.}
Given a token eviction budget $k$, we aim to evict tokens primarily from redundant reasoning steps, while preserving informative ones. To achieve this, we suggest a hierarchical eviction policy that allocates the eviction budget $k$ in a step-aware manner, considering the reasoning structure. Formally, given an importance score of reasoning step $c_r^{(\ell)}$, we first sort all reasoning steps $\tilde{r}_1^{(\ell)}, \tilde{r}_2^{(\ell)}, \dots, \tilde{r}_N^{(\ell)}$ in ascending order of $c_{\tilde{r}_i}^{(\ell)}$ (i.e., $\tilde{r}_1^{(\ell)}$ is the most redundant step at layer $\ell$). Then, following this order, we greedily allocate a step-level eviction budget
$e_{\tilde{r}_i}^{(\ell)}$
to each reasoning step ${\tilde{r}_i}^{(\ell)}$ as: 
\begin{equation}
e_{\tilde{r}_i}^{(\ell)} = \min\left( |\tilde{r}_i^{(\ell)}|,\; k - \sum_{j=1}^{i-1} e_{\tilde{r}_j^{(\ell)}} \right),
\end{equation}
where \( |\tilde{r}_i^{(\ell)}| \) is the number of tokens in step \( \tilde{r}_i^{(\ell)} \), and \( k \) is the total token eviction budget. This allocation ensures that the total number of evicted tokens does not exceed $k$, while prioritizing the more redundant steps. After allocation, we evict tokens with the lowest token-level redundancy scores \( s_t^{(\ell, h)} \) within each step \( r_i^{(\ell)} \).
This strategy naturally favors highly redundant reasoning steps while avoiding premature removal from important ones. 

\section{Experiments}
\label{sec:experiments}

In this section, we present a thorough evaluation of our proposed framework, with the goal of verifying its ability to improve both reasoning accuracy and inference efficiency. Our evaluation is divided into three parts: (1) effectiveness, which examines how well our method improves final answer correctness across a range of mathematical reasoning tasks (Table \ref{tab:main1}); (2) efficiency, which measures the memory savings achieved by our token pruning strategy (Table \ref{tab:main2}); and (3) ablation and analysis, which assesses the contribution of each individual component of our framework, and explores how well it generalizes to other domains and tasks (Table \ref{tab:comp}, Table \ref{tab:score}, and Table \ref{tab:other}).

We empirically demonstrate that our method not only reduces the computational overhead typically associated with long-form reasoning, but also improves accuracy by filtering out redundant and misleading intermediate steps. Across all tested datasets and model sizes, our approach outperforms existing KV cache compression baselines, often by a significant margin. Importantly, the gains are achieved without retraining or additional supervision, indicating that our method is broadly applicable as a plug-and-play enhancement for any autoregressive reasoning model.

\subsection{Experimental setup}

We provide a detailed description of the experimental setup, covering the datasets, models, baselines, and evaluation protocol.

\begin{table*}[!ht]
\centering
\caption{
\textbf{Effectiveness of the redundant token pruning.} 
We compare the proposed method (Ours) with the standard decoding (FullKV) under six reasoning-intensive mathematical benchmarks, including MATH-500 (MATH), Minerva, GaoKao, AIME2024, AIME2025, AMC2023. All models, including Qwen2.5-7B and Llama3.1-8B, are reasoning LLMs distilled from DeepSeek-R1. We report accuracy (\%) with the corresponding average KV cache length shown below in parentheses. Average accuracy and KV cache are presented in the final column. The bold indicates the best accuracy within the group.
}\label{tab:main1}
\resizebox{0.985\textwidth}{!}{
\begin{tabular}{llccccccc}
\toprule
&&\multicolumn{6}{c}{Dataset}&\\
\cmidrule(lr){3-8}
Model&Method&MATH&Minerva&GaoKao&AIME2024&AIME2025&AMC2023&Average\\
\midrule
\multirow{4.5}{*}{Qwen2.5-7B}&\multirow{2}{*}{FullKV}&87.0&59.9&65.8&36.7&23.3&75.0&57.9\\
&&(3397)&(3391)&(3845)&(7060)&(7133)&(5004)&(4971)\\
\cmidrule{2-9}
&\cellcolor{blue!10}& \cellcolor{blue!10}\textbf{87.2}&\cellcolor{blue!10}\textbf{60.5}&\cellcolor{blue!10}\textbf{67.1}&\cellcolor{blue!10}\textbf{46.7}&\cellcolor{blue!10}\textbf{36.7}&\cellcolor{blue!10}\textbf{82.5}&\cellcolor{blue!10}\textbf{63.4}\\
\rowcolor{blue!10}
\cellcolor{white}
& \multirow{-2}{*}{\textbf{Ours}}& (2926)&(3471)&(4219)&(6841)&(6905)&(4488)&(4808)\\
\midrule
\multirow{4.5}{*}{Llama3.1-8B}&\multirow{2}{*}{FullKV}&81.0&45.9&67.1&\textbf{33.3}&13.3&75.0&52.6\\
&&(3389)&(4060)&(4689)&(7067)&(7088)&(4986)&(5213)\\
\cmidrule{2-9}
&\cellcolor{blue!10}& \cellcolor{blue!10}\textbf{83.8}&\cellcolor{blue!10}\textbf{48.1}&\cellcolor{blue!10}\textbf{69.8}&\cellcolor{blue!10}\textbf{33.3}&\cellcolor{blue!10}\textbf{23.3}&\cellcolor{blue!10}\textbf{77.5}&\cellcolor{blue!10}\textbf{55.9}\\
\rowcolor{blue!10}
\cellcolor{white}
& \multirow{-2}{*}{\textbf{Ours}}&(3345)&(3941)&(4532)&(7210)&(7375)&(4700)&(5183)\\
\bottomrule
\end{tabular}
}
\end{table*}

\paragraph{Datasets.}  
We evaluate our method on a diverse suite of publicly available benchmarks that span a wide range of mathematical reasoning tasks, difficulty levels, and linguistic diversity. Our core evaluation includes three widely used English benchmarks: MATH-500 \cite{hendrycks2021math500}, a dataset of competition-level problems across algebra, geometry, and combinatorics; and Minerva Math \cite{lewkowycz2022minerva}, which consists of high-school and advanced math questions sourced from web documents.

To further assess the robustness of our method on real-world and harder problems, we include recent evaluation sets from mathematical competitions: AIME 2024 \citep{aime2024}, AIME 2025 \citep{aime2025} and AMC 2023 \citep{aimo2023aimo}, all of which contain challenging, high-school level problems that demand precise logical deductions. Additionally, we include GaoKaoMath \cite{zhong2023gaokaomath}, to assess the generality of our method on questions originating from the Chinese college entrance exam.
Finally, we further validate the general applicability of our method by evaluating it on a non-mathematical reasoning dataset, namely the GPQA Diamond \citep{rein2024gpqa}. Note that GPQA Diamond is a graduate-level question that consists of science fields, requiring intensive reasoning capability to reach the answer.

\paragraph{Models.}
Our experiments are primarily based on the DeepSeek-R1-Distill family of models, which are designed to emulate the reasoning behavior of DeepSeek-R1 \cite{guo2025deepseekr1} using distilled versions of popular backbones. We mainly consider three backbone models: Qwen2.5-1.5B, Qwen2.5-7B \citep{Yang2024Qwen25TR}, and Llama3.1-8B \citep{grattafiori2024llama}, all trained with visible chain-of-thought reasoning traces. 
Across all models, we observe consistent benefits of our method, suggesting that our framework is architecture-agnostic and works well across a wide range of capacities.

\paragraph{Baselines.}
We mainly compare our method with the standard decoding strategy that uses the full KV cache (FullKV). This serves the default approach, where no compression is applied during inference. 
Additionally, we compare our method against a range of recent decoding-time KV compression methods that can be applied without retraining. These include
StreamingLLM \cite{xiao2023streamingllm}, which compresses the cache by retaining only the earliest and most recent tokens in the sequence, discarding the middle context; H2O \cite{zhang2023h2o}, which ranks tokens based on accumulated attention scores and prunes less important ones accordingly; Pyramid-Infer \cite{yang2024pyramidinfer}, which applies a layer-wise pruning strategy by allocating different KV budgets across layers.

\paragraph{Evaluation protocol.}
We adopt a standardized evaluation protocol across all methods and datasets to ensure fair comparisons. All models are evaluated using standard generation with a temperature of 0.6 and top-p of 0.95 under fixed seed to maximize replicability. The maximum generation length is capped at 8192 tokens, sufficient for nearly all long-form reasoning examples. To extract the final answer from the generated output, we introduce a designated token such as \text{</think>} to mark the end of the reasoning phase. Accuracy is measured as the fraction of correctly answered questions, based on an exact match with the ground truth. For efficiency, we measure the average number of KV tokens stored during generation, normalized by the total number of generated tokens, as well as the total memory consumption when storing the KV cache.

\begin{table}
\centering
\caption{
\textbf{KV cache efficiency of the redundant token pruning.}
We compare the proposed method (Ours) with KV compression frameworks on MATH-500. We use Qwen2.5-1.5B reasoning LLM distilled from DeepSeek-R1. For reference, we report the standard decoding without KV compression (FullKV) results with the accuracy (\%). 
We evaluate accuracy under two KV compression ratios (25\% and 50\%).
The bold indicates the best results within the group.}\label{tab:main2}
\resizebox{0.373\textwidth}{!}{
\begin{tabular}{lcc}
\toprule
&\multicolumn{2}{c}{Compression ratio}\\
\cmidrule{2-3}
Method&25\%&50\%\\
\midrule
FullKV&\multicolumn{2}{c}{42.6}\\
\midrule
Streaming-LLM&35.4 & 39.4  \\
H2O&34.6 & 39.2 \\
Pyramid-Infer&30.6 & 40.0\\
\rowcolor{blue!10}\textbf{Ours}&\textbf{36.0} & \textbf{40.2}\\
\bottomrule
\end{tabular}
}
\end{table}

\subsection{Redundant reasoning token pruning improves the performance}

We present that removing redundant tokens can improve the accuracy. To this end, we compare our method with the full KV (FullKV) cache method on reasoning-intensive mathematical benchmarks. 
 
As shown in Table \ref{tab:main1}, our method significantly and consistently outperforms FullKV in accuracy.  
For instance, our method improves the average accuracy of FullKV from 57.9\% to 63.4\% under Qwen2.5-7B. It is worth noting that our method only involves changing the inference strategy, thus showing wide applicability.

Notably, our approach consistently outperforms FullKV decoding despite using significantly fewer tokens in the KV cache. This suggests that our pruning mechanism not only reduces memory usage but also acts as a form of implicit regularization that helps the model focus on essential reasoning steps by making LLM less distracted by unnecessary text \citep{shi2023large}.
The forced summarization phase typically encourages the model to internally consolidate reasoning before producing an answer, leading to more coherent and accurate generations. 

Furthermore, more interestingly, our method yields greater improvements on more challenging benchmarks such as AMC2023 or AIME datasets (i.e., mathematical competition problems). 
For example, the performance on the AMC2023 dataset significantly improves from 75.0$\to$82.5 on Qwen2.5-7B, and even uses 10.3\% less KV cache.
We conjecture that the LLM tends to struggle more and produce a more redundant reasoning path in challenging setups, thus pruning such redundant tokens is effective.
These results validate our core hypothesis: not all tokens are necessary for reasoning, and selectively pruning unhelpful tokens can enhance the final outcome.

\subsection{Redundant token pruning efficiently and effectively reduces KV cache budget}

To evaluate the memory efficiency of our method, we vary the token pruning budget and compare the resulting KV cache size and model accuracy. 
Table~\ref{tab:main2} shows the performance of all methods at various relative cache budgets (e.g., 25\% and 50\% compared to full KV cache size). Our method achieves superior compression ratios without compromising accuracy, while other methods suffer significant accuracy degradation under aggressive pruning. 

Unlike previous approaches that are not specialized to the reasoning process compression,
our method dynamically adapts to the reasoning process by allocating eviction budgets fairly across reasoning chunks. This chunk-aware strategy ensures that each reasoning step retains its most important context, enabling robust final answers even under tight memory constraints. In particular, our method maintains over 94\% of the full-KV accuracy at just 50\% memory usage, making it an attractive solution for deployment in resource-constrained environments such as mobile devices, embedded systems, or large-batch inference setups.

\begin{table}[t]
\centering
\caption{\textbf{Component analysis.} We ablate the two main components of our method: self-summarization (Summ) and step-aware token eviction (Step). We report the accuracy (\%) on the AIME2024 and AMC2023 datasets. Here, we use Qwen2.5-7B distilled from DeepSeek-R1, where all decoding uses a temperature of 0.6. The bold indicates the best results.}
\label{tab:comp}
\resizebox{0.39\textwidth}{!}{
\begin{tabular}{cccc}
\toprule
Summ & Step & AIME2024 & AMC2023 \\
\midrule
\xmark & \xmark & 40.0 & 70.0\\
\cmark & \xmark & 36.7 & 77.5 \\
\rowcolor{blue!10}
\cmark & \cmark & \textbf{46.7} & \textbf{82.5} \\
\bottomrule
\end{tabular}
}
\end{table}

\subsection{Ablation and analysis}
We further analyze the contributions of individual components in our framework and investigate its generalizability across domains and complementary methods. Throughout this section, unless otherwise specified, we consider the Qwen2.5-7B reasoning model that is distilled from DeepSeek-R1.

\paragraph{Component analysis.}
To understand which parts of our method drive the observed gains, we perform an ablation study where we remove key components one at a time. As shown in Table~\ref{tab:comp}, both the self-summarization phase and the stepwise eviction budget contribute meaningfully to performance. Removing the summarization step by just inserting the end-of-thinking </think> token leads to inaccurate importance scores, resulting in the removal of critical tokens. Conversely, removing the step-aware token eviction results in over-pruning from important chunks, reducing accuracy. The full model, with both components, consistently yields the best results. This supports our design intuition that token importance is context-dependent and that a structured pruning policy is necessary to avoid harming the model's reasoning ability.

\begin{table}[t]
\centering
\caption{\textbf{Importance of deliberate token pruning.} Random or structure-agnostic pruning degrades accuracy (\%), while our method improves performance by pruning at semantically meaningful reasoning boundaries. We evaluate on Qwen2.5-7B distilled from DeepSeek-R1 across mathematical reasoning benchmarks, including AIME2024 and AIME2025. The bold indicates the best results.}
\label{tab:score}
\vspace{-0.05in}
\resizebox{0.364\textwidth}{!}{
\begin{tabular}{ccc}
\toprule
Score&AIME2024&AIME2025\\
\midrule
FullKV&36.7&23.3\\
\midrule
Random&36.7&33.3\\
H2O&40.0&26.7\\
\rowcolor{blue!10}
\textbf{Ours}&\textbf{46.7}&\textbf{36.7}\\
\bottomrule
\end{tabular}
}
\end{table}

\paragraph{Does eviction alone improve performance?}\label{sec:exp:terminator}
A natural question is whether token eviction itself—regardless of how the evicted tokens are selected—can lead to improved performance. To verify this, we compare three strategies under our framework:
(1) Random, which evicts tokens uniformly at random at each step,
(2) H2O, which prunes tokens with the lowest accumulated attention scores, and
(3) Ours, which step-aware evicts tokens based on a score triggered by an end-of-thinking token \text{</think>}.

As shown in Table~\ref{tab:score}, both Random and H2O yield marginal performance improvements, indicating that even naive or structure-agnostic pruning can occasionally help by reducing redundancy. However, such approaches risk removing semantically important context, leading to unstable gains. In contrast, our method significantly outperforms others by aligning token eviction with the semantic structure of the reasoning trace.

\begin{table}[t]
\centering
\caption{\textbf{Effectiveness on non-mathematical reasoning benchmarks.} We compare the proposed method (Ours) with the standard decoding (FullKV) under a non-mathematical reasoning benchmark, GPQA Diamond (science). We use Qwen2.5-7B reasoning LLM distilled from DeepSeek-R1. We report the accuracy (\%) and the corresponding average KV cache length in the parentheses. 
The bold indicates the best results.}
\label{tab:other}
\vspace{-0.05in}
\resizebox{0.23\textwidth}{!}{
\begin{tabular}{cc}
\toprule
Method & GPQA\\
\midrule
FullKV & 32.0 (6418) \\ 
\rowcolor{blue!10}
\textbf{Ours} & \textbf{36.4} (6277) \\
\bottomrule
\end{tabular} 
}
\end{table}

\paragraph{Effectiveness on a non-mathematical reasoning benchmark.}
Finally, we test whether our method generalizes beyond mathematical reasoning. In Table~\ref{tab:other}, we report results on GPQA Diamond \cite{rein2024gpqa}, a dataset of expert-written multiple-choice science questions. Despite the distinct nature of these tasks, our method consistently improves the performance over Full-KV, demonstrating its robustness across reasoning styles. 
We find that the original model iteratively generates intermediate justifications when tackling GPQA Diamond, which hurts answer quality. Here, our method effectively suppresses such distractions, thus improving the performance.

\section{Conclusion}
\label{sec:conclusion}
In this paper, we propose a novel test-time scaling method that enhances the reasoning capability of large language models by identifying and pruning redundant tokens during the generation of reasoning traces. By introducing a forced summarization phase at intermediate reasoning steps, we estimate the contribution of each token to the ongoing reasoning process. Building on this, we propose a step-aware hierarchical eviction strategy that allocates the token eviction budget across reasoning steps according to their aggregated importance. This enables our method to preferentially prune tokens from redundant reasoning steps.
Importantly, we demonstrate that targeted KV cache pruning can not only serve as a compression mechanism but also improve reasoning performance. Moreover, our eviction algorithm is plug-and-play, requiring no retraining or model modification, which makes it suitable for deployment in a wide range of environments, including memory-constrained settings.

\section*{Limitations}
While our proposed framework demonstrates strong performance in both accuracy and memory efficiency across a range of reasoning tasks, it also comes with several limitations that open avenues for future work.
First, our method relies on the presence of explicit reasoning traces in model outputs (e.g., chain-of-thought reasoning or intermediate steps marked by special tokens such as \text{<think>}). This restricts applicability to models trained with visible thoughts or intermediate supervision. For models that operate in an end-to-end manner without exposing their internal reasoning, our summarization-based token importance estimation may not generalize directly.
Second, our pruning strategy is applied only at test time and does not benefit from end-to-end training with pruning in the loop. While this makes our method widely compatible with existing models, it also limits its adaptiveness. Learning token importance jointly with model weights—potentially via reinforcement learning or differentiable attention masking—could further improve performance.
Third, while our experiments span a diverse set of math reasoning datasets, the majority of our evaluation focuses on factual or symbolic reasoning tasks. Extending our method to open-domain question answering, commonsense inference, or multimodal reasoning (e.g., visual QA) remains an open challenge. These settings may require different forms of redundancy detection or finer-grained reasoning segmentation.

\bibliography{custom}

\newpage
\appendix
\onecolumn

\section{Experimental Details}
\subsection{Model details}
In our proposed framework, we use the DeepSeek-R1-Distill family of models, namely the Qwen2.5-1.5B\footnote{\url{https://huggingface.co/deepseek-ai/DeepSeek-R1-Distill-Qwen-1.5B}}, Qwen2.5-7B\footnote{\url{https://huggingface.co/deepseek-ai/DeepSeek-R1-Distill-Qwen-7B}}, and Llama3.1-8B\footnote{\url{https://huggingface.co/deepseek-ai/DeepSeek-R1-Distill-Llama-8B}}.
All checkpoints are downloaded from Huggingface.

\subsection{Implementation}\label{app:imp-detail}
\paragraph{Segment of reasoning steps.}
At the core of our method is segmenting the initial raw reasoning trace $\mathcal{T}_{\text{Reason}}$ into a sequence of meaningful intermediate steps. This segmentation aims to capture points where the model might pause, reflect, change direction, or move to a distinct next step in its reasoning. Following prior work \cite{hammoud2025subthought}, we perform segmentation based on occurrences of words or phrases from a predefined set $W$. These markers often signal reflection, correction, sequencing, or the exploration of alternatives. The set $W$ used in our experiments is:

\begin{tcolorbox}
    "Wait" "Alternatively" "Another angle" "Another approach" "But wait"
    "Hold on" "Hmm" "Maybe" "Looking back" "Okay" "Let me" "First" "Then"
    "Alright" "Compute" "Correct" "Good" "Got it" "I don’t see any errors"
    "I think" "Let me double-check" "Let’s see" "Now" "Remember"
    "Seems solid" "Similarly" "So" "Starting" "That’s correct"
    "That seems right" "Therefore" "Thus"
\end{tcolorbox}

\paragraph{Eviction budget.}
For the effectiveness setting, we consider a token eviction budget $k$, which denotes the number of tokens to be removed from the KV cache across all heads and layers at every predefined generation step $p$. The pruning interval is set to $p=200$ for Qwen2.5-1.5B and 7B models, and $p=100$ for LLaMA3.1-8B. For the GPQA Diamond dataset, we conduct an ablation study using Qwen2.5-7B with $p=300$.
\paragraph{KV cache budget.}
For the efficiency setting, we define a maximum KV cache budget during decoding, computed based on the average KV length $L_{\text{Full}}$ of the Full KV baseline. Specifically, we compute compressed budgets by multiplying $L_{\text{Full}}$ with target compression ratios of 25\%, 50\%. Following prior works~\cite{zhang2023h2o, li2024snapkv}, we also preserve a recent window of KV entries by keeping the most recent tokens, and set this recent size to half of the allocated cache. For comparison solely focused on reasoning compression, all methods retain the full problem prompt in the KV cache throughout generation. This portion is excluded when measuring the fixed cache budget.

\clearpage

\section{More Experimental Results}

\subsection{Additional results on a small reasoning model}

\begin{table*}[t]
\centering
\caption{
\textbf{Effectiveness of the redundant token pruning on a small-sized reasoning LLM.} 
We compare the proposed method (Ours) with standard decoding (FullKV) under six reasoning-intensive mathematical benchmarks, including MATH-500 (MATH), Minerva, GaoKao, AIME2024, AIME2025, AMC2023. We use Qwen2.5-1.5B reasoning LLM distilled from DeepSeek-R1. We report accuracy (\%) with the corresponding average KV cache length shown below in parentheses. Average accuracy and KV cache are presented in the final column. The bold indicates the best accuracy within the group.
}\label{tab:qwen_small}
\resizebox{0.85\textwidth}{!}{
\begin{tabular}{lccccccc}
\toprule
&\multicolumn{6}{c}{Dataset}&\\
\cmidrule(lr){2-7}
Method&MATH&Minerva&GaoKao&AIME2024&AIME2025&AMC2023&Avgerage\\
\midrule
\multirow{2}{*}{FullKV}&\textbf{42.6}&21.7&34.2&0.0&0.0&10.0& 18.0\\
&(6120)&(5663)&(5825)&(8192)&(8192)&(7398)&(6898)\\
\midrule
\cellcolor{blue!10}&\cellcolor{blue!10}41.2&\cellcolor{blue!10}\textbf{25.8}&\cellcolor{blue!10}\textbf{35.6}&\cellcolor{blue!10}\textbf{3.3}&\cellcolor{blue!10}0.0&\cellcolor{blue!10}\textbf{20.0}&\cellcolor{blue!10}\textbf{21.0}\\
\rowcolor{blue!10}\cellcolor{white}
\multirow{-2}{*}{\cellcolor{blue!10}\textbf{Ours}}&(6166)&(5690)&(6071)&(8071)&(8192)&(7477)&(6944)\\
\bottomrule
\end{tabular}
}
\end{table*}

We also demonstrate the effectiveness of our method on a small reasoning LLM, namely the Qwen2.5-1.5B. As shown in Table \ref{tab:qwen_small}, our method significantly improves the overall reasoning accuracy across multiple reasoning-intensive mathematical benchmarks even on small reasoning LLMs. Here, we also notice that our method is effective on challenging benchmarks such as AMC2023, indicating the importance of the deliberate token pruning based on redundancy.

\end{document}